\documentclass[letterpaper, 10 pt, conference]{ieeeconf}  

\IEEEoverridecommandlockouts                              

\usepackage{graphics} 
\usepackage{url}
\usepackage{hyperref}
\usepackage{graphicx}
\usepackage{subcaption}

\title{\LARGE \bf
EdgeVLA: Efficient Vision-Language-Action Models
}

\author{Paweł Budzianowski, Wesley Maa, Matthew Freed, Jingxiang Mo, Winston Hsiao \\ Aaron Xie, Tomasz Młoduchowski, Viraj Tipnis, Benjamin Bolte 
\thanks{Paweł Budzianowski, Wesley Maa,  Matthew Freed, Winston Hsiao, Aaron Xie, Tomasz Młoduchowski, Viraj Tipnis, Benjamin Bolte are with K-Scale Labs. Email correspondence to  {\tt\small pawel@kscale.dev}}
\thanks{Jingxiang Mo is with McGill University.}
}

\begin{document}

\maketitle
\thispagestyle{empty}
\pagestyle{empty}

\begin{abstract}
Vision-Language Models (VLMs) have emerged as a promising approach to address the data scarcity challenge in robotics, enabling the development of generalizable visuomotor control policies. While models like OpenVLA showcase the potential of this paradigm, deploying large-scale VLMs on resource-constrained mobile manipulation systems remains a significant hurdle. This paper introduces Edge VLA (EVLA), a novel approach designed to significantly enhance the inference speed of Vision-Language-Action (VLA) models. EVLA maintains the representational power of these models while enabling real-time performance on edge devices. We achieve this through two key innovations: 1) Eliminating the autoregressive requirement for end-effector position prediction, leading to a 7x speedup in inference, and 2) Leveraging the efficiency of Small Language Models (SLMs), demonstrating comparable training performance to larger models with significantly reduced computational demands. Our early results demonstrate that EVLA achieves comparable training characteristics to OpenVLA while offering substantial gains in inference speed and memory efficiency. We release our model checkpoints and training \href{https://github.com/kscalelabs/evla }{codebase} to foster further research. 
\end{abstract}

\section{INTRODUCTION}
\label{sec:intro}
The development of robust and generalizable manipulation policies has long been hampered by the limited availability of large-scale, diverse embodied datasets. Recent advancements in Vision-Language Models (VLMs) \cite{llava, prismatic} offer a compelling solution to this challenge. By leveraging the vast amount of readily available image-text data, VLMs can learn rich representations of the world and be adapted for visuomotor control tasks. Open-source models like OpenVLA \cite{openvla} have demonstrated the effectiveness of this approach, showcasing impressive performance in various robotic manipulation tasks.
However, deploying these large-scale VLMs, often exceeding billions of parameters, on resource-constrained mobile platforms with edge devices like the Jetson Nano presents significant challenges. Their high computational and memory requirements hinder real-time performance and limit accessibility for researchers and practitioners. 

The progress in the mobile manipulation can be effective only if the systems we design are inexpensive and easily deployable without putting too much strain on compute requirements. That is why, this paper introduces Edge VLA (EVLA), a novel VLA architecture designed to address above-mentioned challenges. EVLA offers significant improvements in inference speed and efficiency without compromising foundation models' representational power. Our approach centers around two key innovations: First, our work focuses on architectural modifications to achieve significant speedups while maintaining model performance. Specifically by eliminating the autoregressive requirement for end-effector prediction and leveraging the efficiency of SLMs. We challenge the autoregressive approach for predicting end-effector positions, demonstrating that joint control, where the entire position is predicted simultaneously, does not diminish the model's encoding capabilities. This modification yields a $7$-times increase in inference speed, crucial for real-time robotic control on edge devices.

Secondly, we explore the potential of recently developed Small Large Language Models (SMLs), such as Qwen2 \cite{qwen2}, Phi \cite{phi3} or Gemma \cite{gemma2}, which achieve comparable performance to their larger counterparts thanks to scaling laws with significantly reduced computational footprints.
Our proposed architecture EVLA comprises of a pretrained language model Qwen2-0.5B fused with two visual encoders SigLIP \cite{siglip} and DINOv2 \cite{dinov2} adding to 1B parameters. EVLA maintains training performance comparable to that of models $7$ times larger while significantly reducing hardware requirements.

\begin{figure*}[ht!] 
  \centering
  \includegraphics[width=\textwidth]{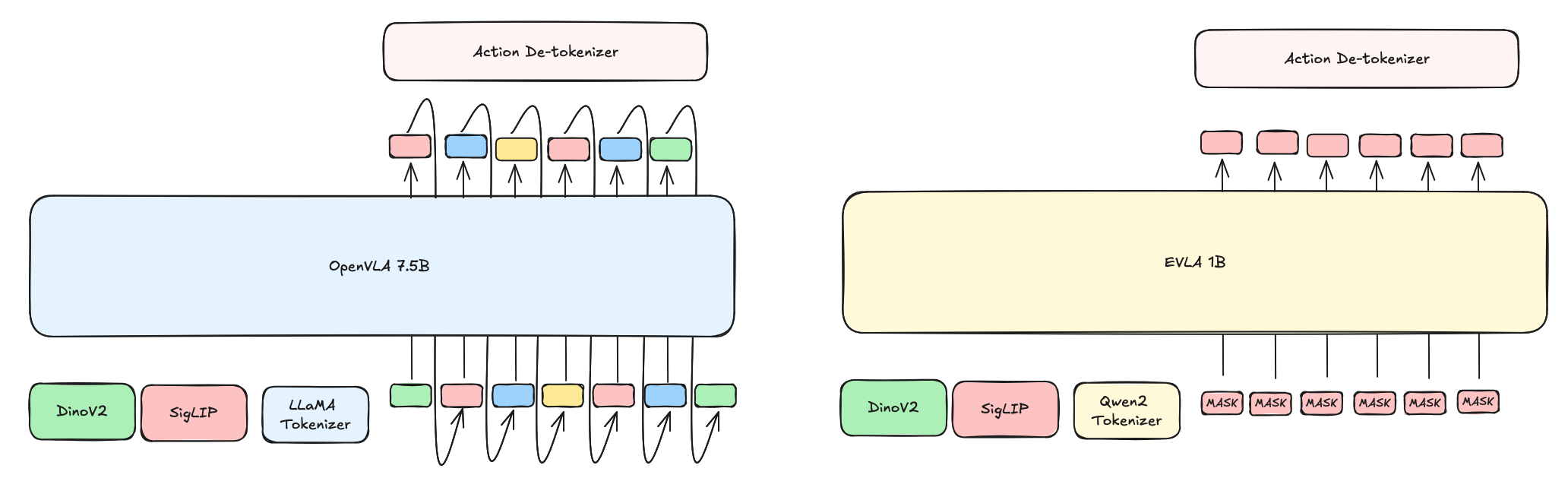} 
  \caption{The comparison of generation logic between OpenVLA and EVLA. The pretraining phase is identical for both models. In phase two, the EVLA LLM is being retrained to generate end-effector position in an autoregressive fashion.}
  \label{fig:architecture}
\end{figure*}
\section{RELATED WORK}
\label{sec:rw}
Learning-based approaches to mobile manipulation are beginning to reach or exceed the performance of classical model-based control systems \cite{mobilealoha}. We can broadly divide these approaches into systems trained from scratch and those fine-tuned on top of the foundation models.

The former approach has relied on behavioral cloning where visual observations are typically mapped to either end effector position with orientation or joint positions \cite{diffusionpolicy, vqbet}. These models can be enhanced through regularization, planning or multi-task learning pushing the limits of the performance. This approach enables easy deployment with relatively cheap hardware but does not leverage the power of foundation models \cite{hellorobot, mobilealoha}. These systems typically train models from scratch with model sizes from $10$ to $100$M parameters which limits their ability to generalize to novel environments \cite{aloha,diffusionpolicy}.

The line of work that relies on foundation models incorporates all the aforementioned techniques while aiming for more powerful generalization capabilities. The most extensively explored approach relies on vision-language models \cite{llava, prismatic}. The vision component is typically adapted to operate in the same token space as the LLM, allowing for the reuse of different pretrained blocks.
Combined with large manipulation datasets such as OpenX \cite{openx}, these models have demonstrated the promise of this paradigm by generalizing to new environments \cite{rt2, openvla}. Although these works have highlighted the potential of leveraging large language models (LLMs), they come with substantial computational demands. Efforts to improve efficiency include quantization techniques \cite{wang2023bitnetscaling1bittransformers} and hardware-specific kernels \cite{triton}. Nevertheless, these system achieve speed of only $5$ to $10$ Hz with stationary compute systems preventing their deployment on edge devices even in laboratory settings.

\begin{figure*}[th!] 
  \centering
  \begin{subfigure}[b]{0.49\textwidth} 
    \includegraphics[width=\textwidth]{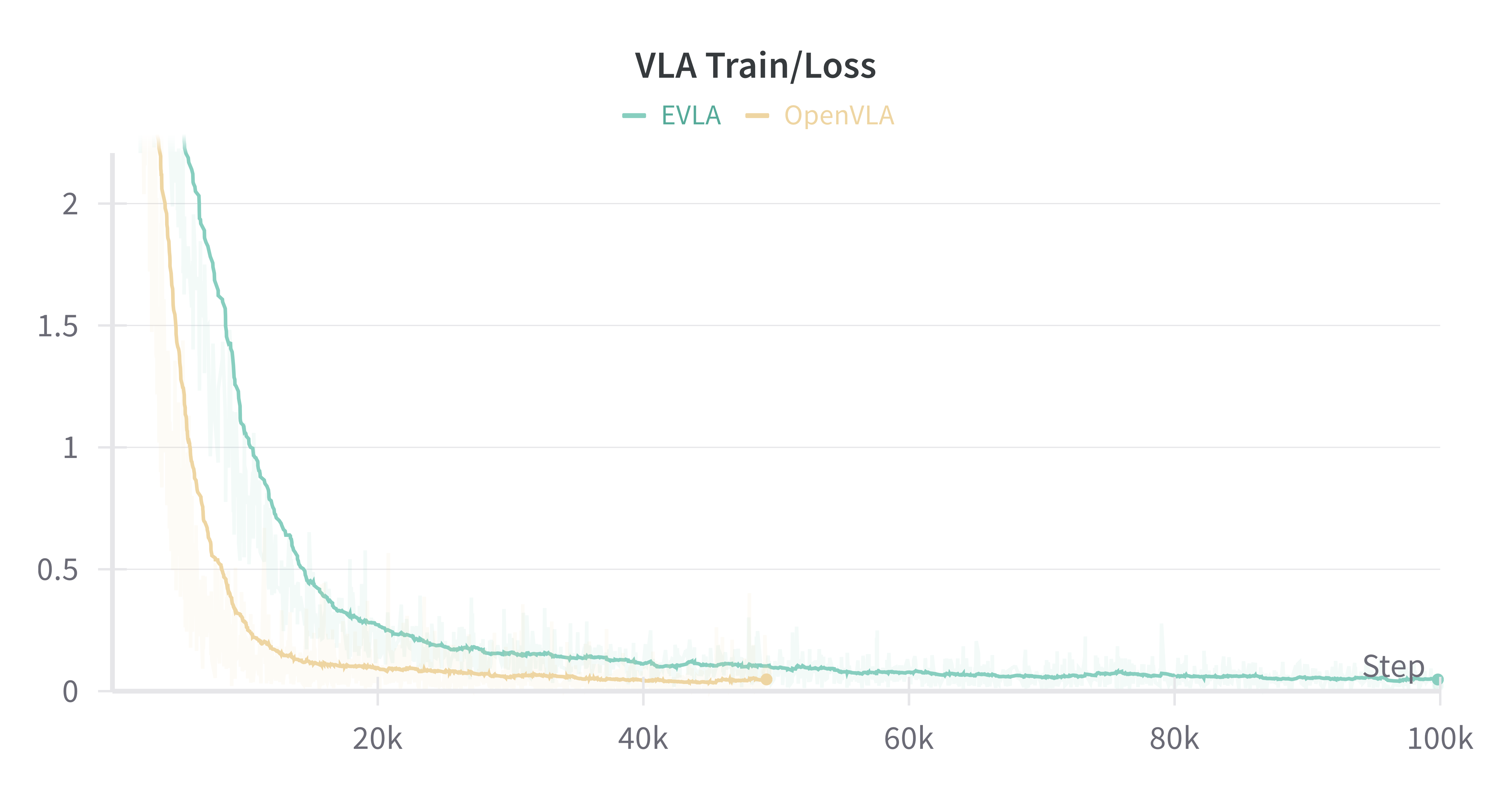}
    \label{fig:bridge_loss}
  \end{subfigure}
  \hfill 
  \begin{subfigure}[b]{0.49\textwidth}
    \includegraphics[width=\textwidth]{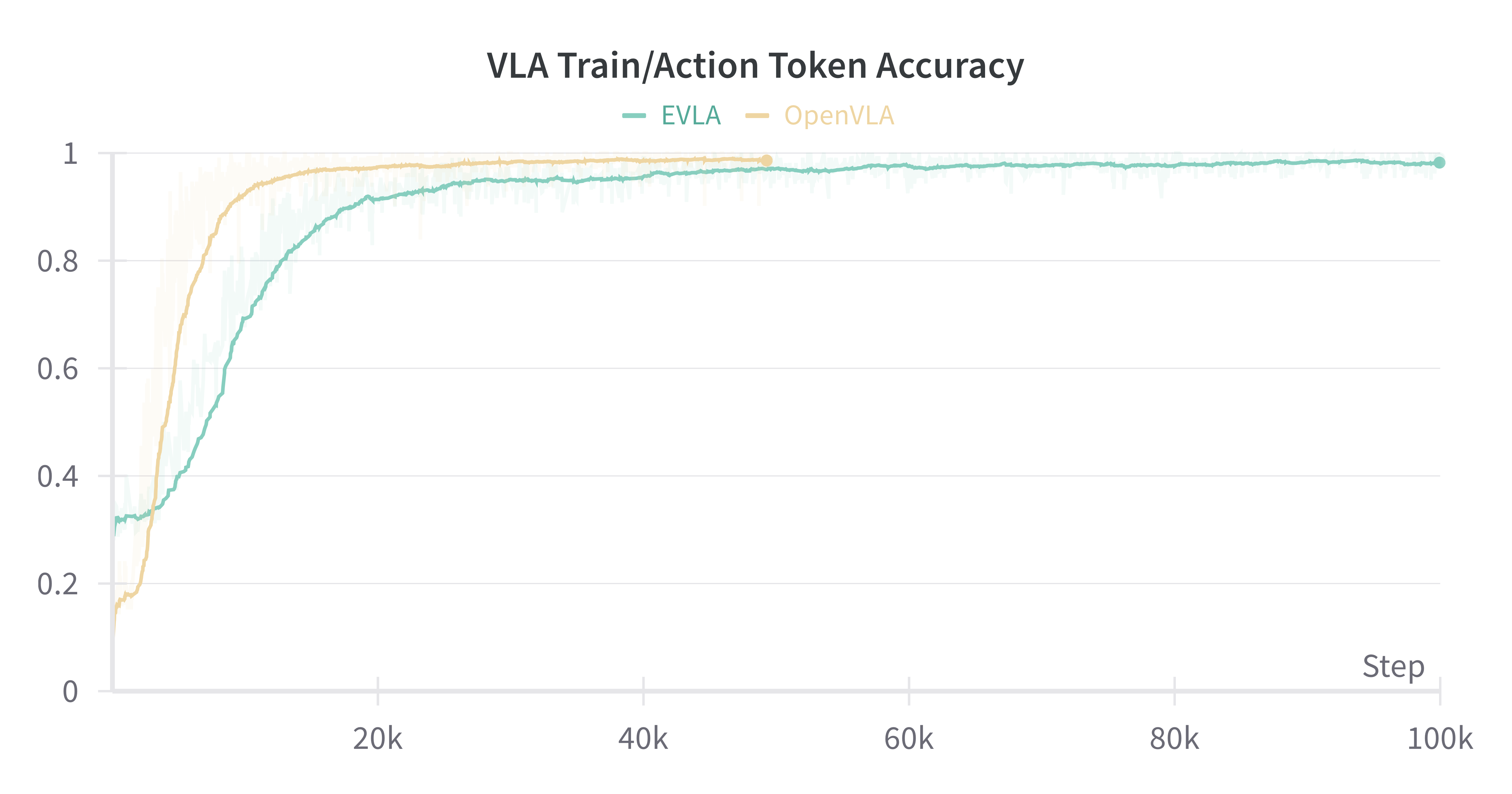}
    \label{fig:subfigure2}
  \end{subfigure}
  \caption{The loss (left) and action token accuracy (right) training curves for both OpenVLA and EVLA models during training on the BridgeData V2 dataset.}
  \label{fig:bridge_acc}
\end{figure*}

\section{METHOD}
\label{sec:method}
\subsection{Phase 1: VLM Pretraining}
EVLA is based on a VLM trained using a combination of image-text pairs sourced from diverse captioning datasets and synthetically generated multimodal instruction-tuning examples \cite{llava}. The pre-training dataset comprises $1.2$M text-image pairs, facilitating the learning of robust visual and language representations following the recipe of the PrismaticVLM family of models \cite{prismatic}. For language processing, we utilize Qwen2 \cite{qwen2} with $0.5$B parameters as it demonstrates the effectiveness of SLMs in achieving comparable performance to that of larger models. We adopt a two-part visual encoder, employing pretrained SigLIP \cite{siglip} and DinoV2 \cite{dinov2} models, following the architecture of OpenVLA \cite{openvla}. A projection layer that maps the visual representation to the language model's token space is learned jointly with the finetuned visual and language components.

\subsection{Phase 2: Joint Control for End-Effector Prediction}
\label{sec:jointcontrol}
The second phase of training utilizes around $1$M of manipulation examples from the OpenX dataset \cite{openx}. Traditional VLAs employ an autoregressive approach to predicting end-effector positions, mimicking the causal nature of language generation. However, we hypothesize that for robotic control, this restriction is not inherently necessary. We propose that predicting the entire end-effector position jointly, rather than sequentially, does not compromise the model's encoding capabilities while significantly improving inference speed. 

By removing the causal mask in the LLM and training the model to output the entire end-effector position at once, we eliminate autoregressive requirements, achieving a six-times speedup in inference - a critical improvement for real-time applications on edge devices.

See Figure \ref{fig:architecture} for the overall layout of the model and the comparison to its autoregressive counterpart.

\section{EARLY RESULTS}
\label{sec:results}
In order to evaluate EVLA's capabilities of adapting to non-autoregressive loss while utilizing SMLs, we used BridgeData V2 \cite{bridge2} and OpenX datasets \cite{openx} as a testbed. We hypothesize that the early training results will shed some light on model characteristics.

\subsection{BridgeData V2 training characteristics}
Initial experiments on the BridgeData V$2$ dataset conducted on a single node with $8$ A$100$-$80$GB GPUs, validate that EVLA can achieve similar training performance to its $7.5$B parameters counterpart. Figure \ref{fig:bridge_acc} illustrates the training progress, showcasing the comparable performance of the two models. It is worth pointing out that the training efficiency is distinguishably slower for EVLA due to smaller parametrization capabilities.

\begin{figure*}[th!] 
  \centering
  \begin{subfigure}[b]{0.49\textwidth} 
    \includegraphics[width=\textwidth]{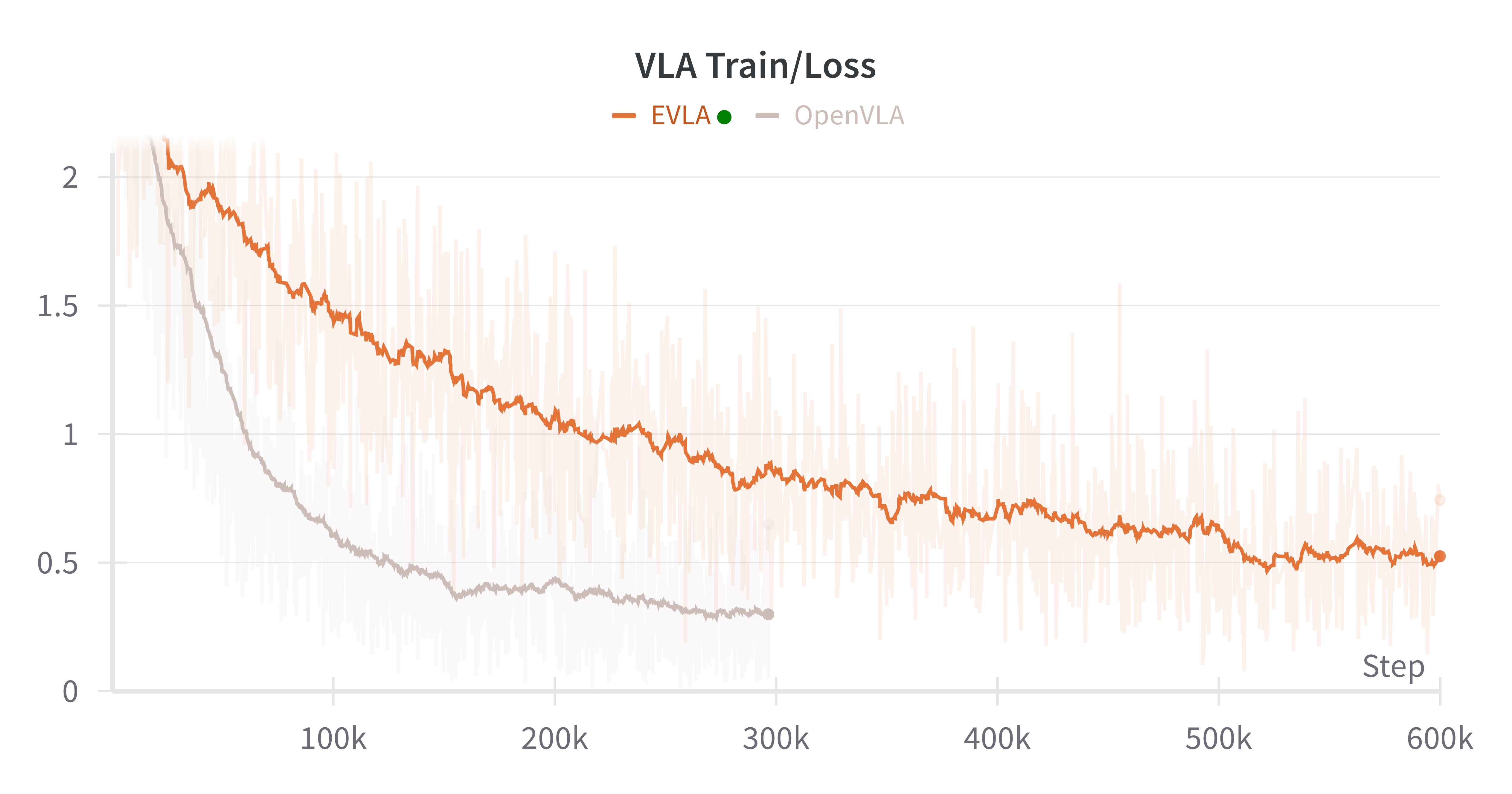}
    \label{fig:oxe_loss}
  \end{subfigure}
  \hfill 
  \begin{subfigure}[b]{0.49\textwidth}
    \includegraphics[width=\textwidth]{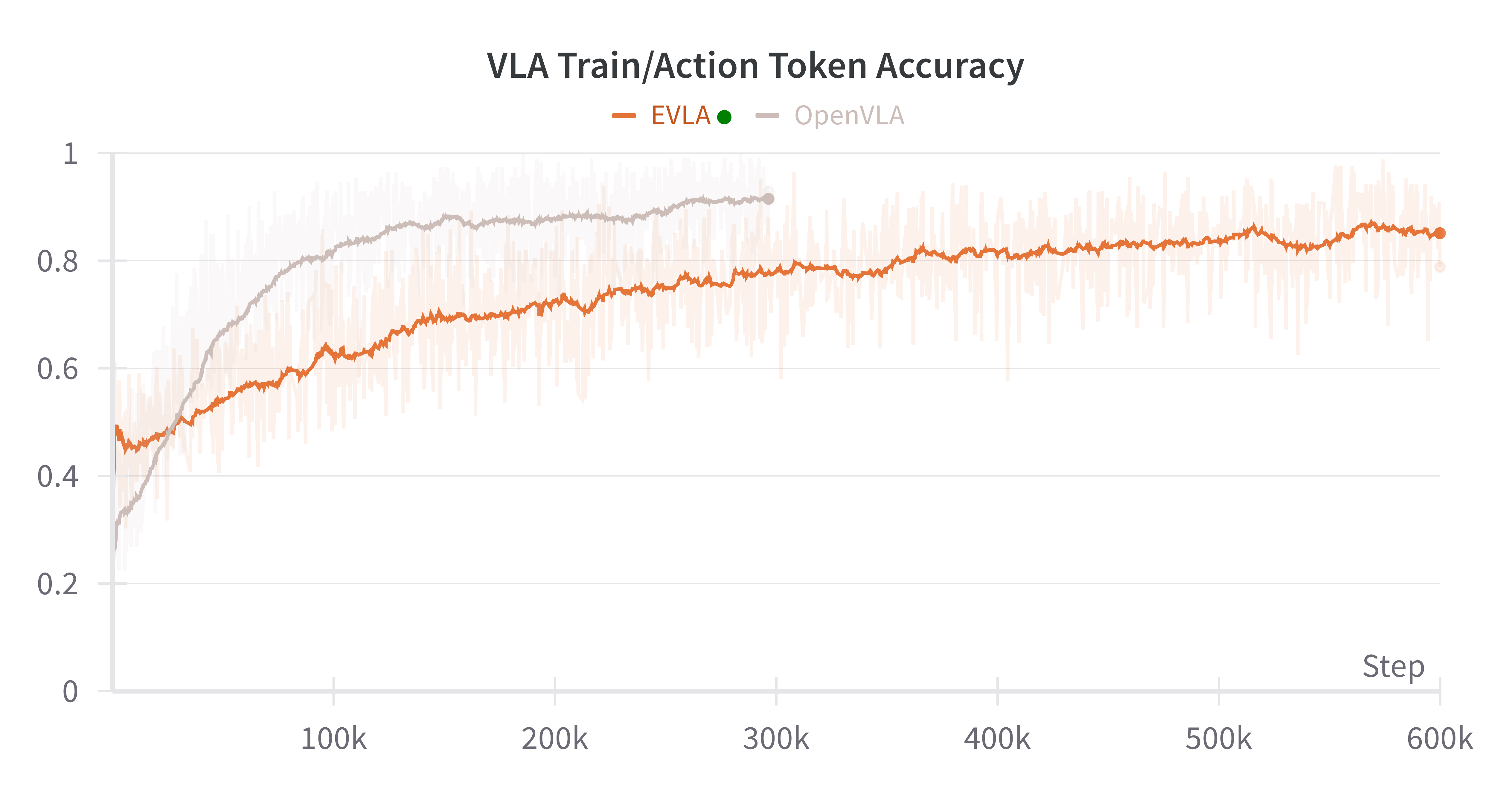}
    \label{fig:oxe_acc}
  \end{subfigure}
  \caption{The loss (left) and the action token accuracy (right) training curves for both OpenVLA and EVLA models during training on the OpenX dataset.}
  \label{fig:oxe_overall}
\end{figure*}

\subsection{OpenX training characteristics}
We further evaluate EVLA on the full OpenX dataset, utilizing 80 A100-40GB GPUs for approximately 5 days. While the training efficiency of EVLA is slower than OpenVLA due to the smaller representational power, the training iteration is around $7$ times faster. It also allows for larger batch sizes, effectively mitigating the difference in training efficiency. Figure \ref{fig:oxe_overall} shows the training progress on the OpenX dataset.

Due to computational constraints, we were not able to reproduce the full training of OpenVLA as in the original implementation \cite{openvla}. However, the training curves show the signs of stagnation and the behavior is similar to the BridgeV2 case.

\subsection{Efficiency Gains}
EVLA's architectural modifications result in substantial improvements in the inference speed and memory consumption, enabling deployment on resource-constrained edge devices. Table \ref{table:efficiency} compares the inference time and memory requirements of EVLA and OpenVLA on an A100-40GB GPU.
\begin{table}[h]
\caption{Efficiency Comparison of EVLA and OpenVLA.}
\label{table:efficiency}
\begin{center}
\begin{tabular}{|c||c|c|}
\hline
Model & Inference Time (ms) & Memory Usage (GB) \\
\hline
OpenVLA & 20 & 16 \\
\hline
EVLA & 5 & 4 \\
\hline
\end{tabular}
\end{center}
\end{table}

By using a smaller VLM and optimizing our architecture, we can achieve significant inference speed and memory improvements. These speedups will only increase with the addition of more degrees of freedom. It is worth noting that OpenVLA uses \verb|flash_attention2| \cite{dao2023flashattention2fasterattentionbetter} kernels, while EVLA is evaluated in the eager mode. Advances in flexible and efficient attention mechanisms, such as FlexAttention \cite{flexattention}, are expected to push these numbers even further. These results show the path for the deployment of mobile manipulation systems on CPU architectures.

\section{CONCLUSIONS}
\label{sec:conclusions}
This paper presents Edge VLA (EVLA), a novel VLA architecture designed for efficient deployment on mobile manipulators and humanoids. By eliminating the autoregressive requirement for end-effector prediction and leveraging the efficiency of SLMs, EVLA achieves significant improvements in inference time and a reduced memory footprint. While the early results suggest EVLA has the potential to be a good candidate for real-time VLA applications on resource-constrained platforms, the crucial next step is to evaluate EVLA on a variety of different embodiments. We plan to employ at least two different humanoid platforms to assess its few-shot capabilities.

We release our model checkpoints and training codebase to facilitate further research. We believe that EVLA's efficiency and accessibility will empower researchers and practitioners to explore the full potential of VLAs for mobile manipulation. Future work will focus on further optimizing EVLA's architecture and exploring its deployment on a wider range of edge devices, including CPU-based platforms.

\addtolength{\textheight}{-12cm}

\bibliographystyle{plain}
\bibliography{refs}

\end{document}